\documentclass[conference]{IEEEtran}
\IEEEoverridecommandlockouts
\usepackage{cite}
\usepackage{amsmath,amssymb,amsfonts}
\usepackage{algorithmic}
\usepackage{graphicx}
\usepackage{textcomp}
\usepackage{lipsum} 
\usepackage{graphicx}
\usepackage{array}
\usepackage[utf8]{inputenc}
\usepackage{makecell} 
\usepackage{comment}

\usepackage[font=small,labelfont=bf,labelsep=period]{caption}
\usepackage{float}
\usepackage{stfloats}

\usepackage{xcolor}
\usepackage{url}
\usepackage{doi}
\def\BibTeX{{\rm B\kern-.05em{\sc i\kern-.025em b}\kern-.08em
    T\kern-.1667em\lower.7ex\hbox{E}\kern-.125emX}}

\begin{document}

\title{eDIF: A European Deep Inference Fabric for Remote Interpretability of LLMs}
\IEEEspecialpapernotice{\footnotesize \textsuperscript{*} Feasibility Study for NDIF-Based AI Interpretability Cluster in Europe}

\author{\mbox{
\begin{tabular}{@{}c c@{}}
Irma Heithoff\textsuperscript{1} & Marc Guggenberger\textsuperscript{1} \\
heithoff22587@hs-ansbach.de & guggenberger22261@hs-ansbach.de \\[0.5em]

Sandra Kalogiannis\textsuperscript{1} & Fabian Maag\textsuperscript{1} \\
s.kalogiannis16258@hs-ansbach.de & f.maag@hs-ansbach.de \\[0.5em]

Susanne Mayer\textsuperscript{1} & Sigurd Schacht\textsuperscript{1,2} \\
s.mayer16905@hs-ansbach.de & sigurd.schacht@hs-ansbach.de \\[0.5em]

Carsten Lanquillon\textsuperscript{2,3} & \\
carsten@coairesearch.org &
\end{tabular}
}}

\maketitle

\begingroup
\renewcommand\thefootnote{}%
\footnotetext{%
  \hspace*{-1.1em}%
  \textsuperscript{1} University of Applied Sciences Ansbach, Ansbach, Germany\\
  \textsuperscript{2} University of Heilbronn, Heilbronn, Germany\\
  \textsuperscript{3} COAI Research, Germany%
}%
\addtocounter{footnote}{-1}%
\endgroup

\begin{abstract}


This paper presents a feasibility study on the deployment of a European Deep Inference Fabric (eDIF), an NDIF-compatible infrastructure designed to support mechanistic interpretability research on large language models. The need for widespread accessibility of LLM interpretability infrastructure in Europe drives this initiative to democratize advanced model analysis capabilities for the research community. The project introduces a GPU-based cluster hosted at Ansbach University of Applied Sciences and interconnected with partner institutions, enabling remote model inspection via the NNsight API. A structured pilot study involving 16 researchers from across Europe evaluated the platform's technical performance, usability, and scientific utility. Users conducted interventions such as activation patching, causal tracing, and representation analysis on models including GPT-2 and DeepSeek-R1-70B. The study revealed a gradual increase in user engagement, stable platform performance throughout, and a positive reception of the remote experimentation capabilities. It also marked the starting point for building a user community around the platform. Identified limitations—such as prolonged download durations for activation data as well as intermittent execution interruptions —are addressed in the roadmap for future development. This initiative marks a significant step towards widespread accessibility of LLM interpretability infrastructure in Europe and lays the groundwork for broader deployment, expanded tooling, and sustained community collaboration in mechanistic interpretability research.
\par\par\par

\end{abstract}

\section{Introduction}

Mechanistic interpretability aims to understand large language models by analyzing their internal components with the goal of uncovering how specific computations arise within a model. This approach has shown promise in identifying concepts responsible for factual recall, syntax, or reasoning in transformer models
\cite{b1, b2}. It complements black-box interpretability by enabling causal interventions and hypothesis-driven analysis \cite{b3}. 

As large-scale models become increasingly central to scientific and commercial applications, the need for transparency and controllability grows. However, access to state-of-the-art systems remains limited. Proprietary APIs from providers like OpenAI or Anthropic restrict inspection of internal activations or gradients, rendering mechanistic research on these systems practically infeasible \cite{b4,b5}. Even open-weight models such as LLaMA 3 or DeepSeek R1 require significant compute resources often unavailable to academic institutions, particularly those without dedicated HPC access.

To address this, infrastructure projects like the National Deep Inference Fabric (NDIF) \cite{b6} have emerged. NDIF enables shared, remote access to large models through a PyTorch-compatible API (NNsight) that supports deferred execution and safe co-tenancy, making detailed experimentation accessible without local deployment.
Beyond its technical design, NDIF explicitly aims to democratize access to interpretability workflows by enabling multiple research groups to share pooled GPU resources and pre-loaded models, thereby lowering the barrier to entry for academic institutions lacking large-scale compute infrastructure.

Building on this concept, this project proposes the European Deep Inference Fabric (eDIF). A scalable, open infrastructure for mechanistic interpretability research based in Europe. In cooperation with the NDIF team, we conducted a feasibility study to explore key questions: Can NDIF-like infrastructure be effectively deployed under European institutional, funding, and regulatory conditions? What technical adaptations and onboarding processes are required? And how well does such a system support real-world interpretability workflows by academic users? Our study evaluates these dimensions through a pilot deployment, with a particular focus on reproducibility, usability, and alignment with European research goals. eDIF aims to lower access barriers, foster collaborative experimentation, and contribute practical insights for building sustainable, open research infrastructure for large-scale LLM analysis in Europe. By doing so, it promotes transparency, reproducibility, and wider societal access to interpretability research in AI.

\section{Related Work
}
A number of open-source tools have emerged to support mechanistic interpretability in transformer models. TransformerLens \cite{b1} enables detailed analysis of internal components, with support for activation patching and causal tracing. Frameworks like pyvene \cite{b7} and baukit \cite{b8} offer complementary abstractions for inspecting and modifying model internals. However, all of these tools require researchers to host models locally and manage large-scale compute environments, a significant barrier when working with frontier models.

To overcome this limitation, NDIF was developed as a remote infrastructure for interpretability research \cite{b6}. NDIF builds on NNsight, a deferred execution API that allows researchers to define interventions in standard PyTorch code, which are then executed remotely on shared, preloaded model instances. This decoupling of experiment design and execution enables co-tenancy, reduces compute costs, and improves accessibility for academic users.

At present, Europe lacks equivalent infrastructure. While supercomputing centers such as LUMI \cite{b9} or MareNostrum 5 \cite{b10} offer general-purpose HPC access, they are not optimized for interactive, model-centric interpretability workflows. Regulatory, funding, and access constraints further limit the practical availability of compute for smaller research groups.

The eDIF project addresses this gap by establishing a European NDIF-compatible cluster designed for mechanistic interpretability. By aligning with existing tooling (e.g., NNsight) and focusing on open-access infrastructure, eDIF supports reproducible, collaborative research within the European context.

\section{System Architecture
}

The eDIF project infrastructure encompasses three computational environments: a primary server at Ansbach University of Applied Sciences, and experimental deployments at Friedrich-Alexander University (FAU) and Heilbronn University, enabling distributed access to neural network introspection capabilities across German research institutions.

\subsection{Hardware and Software Components
}

As part of the feasibility study, the server infrastructure was configured with eight NVIDIA RTX A6000 GPUs, each offering 48 GB of GDDR6 memory \cite{b11}. This high-memory setup was selected to support concurrent deployment of multiple large language model instances per GPU, which is particularly beneficial in the current NDIF framework where efficient GPU sharing across users is limited by per-card memory constraints. Choosing fewer GPUs with larger VRAM thus proved to be a strategically advantageous decision to enable co-tenancy and to assess long-term scalability within institutional compute environments.

The software stack builds upon Ubuntu LTS 22.04, leveraging CUDA for GPU-accelerated computing \cite{b12, b13}. The containerized architecture employs Docker and Docker Compose for application isolation, enabling reliable cross-system deployment and simplified scaling. Ray serves as the core orchestration framework, managing request distribution through queue-based resource allocation \cite{b14}. MinIO provides object storage optimized for AI workloads, supporting efficient large-scale data management \cite{b15}.
The monitoring and analysis infrastructure comprises Loki \cite{b16} for centralized log aggregation, Prometheus \cite{b17} for time-series monitoring, InfluxDB \cite{b18} as a real-time analytics database, and Grafana \cite{b19} as the visualization and monitoring front end.

\subsection{System Architecture and Data Flow}

\begin{figure}
\centerline{\includegraphics[width=0.50\textwidth]{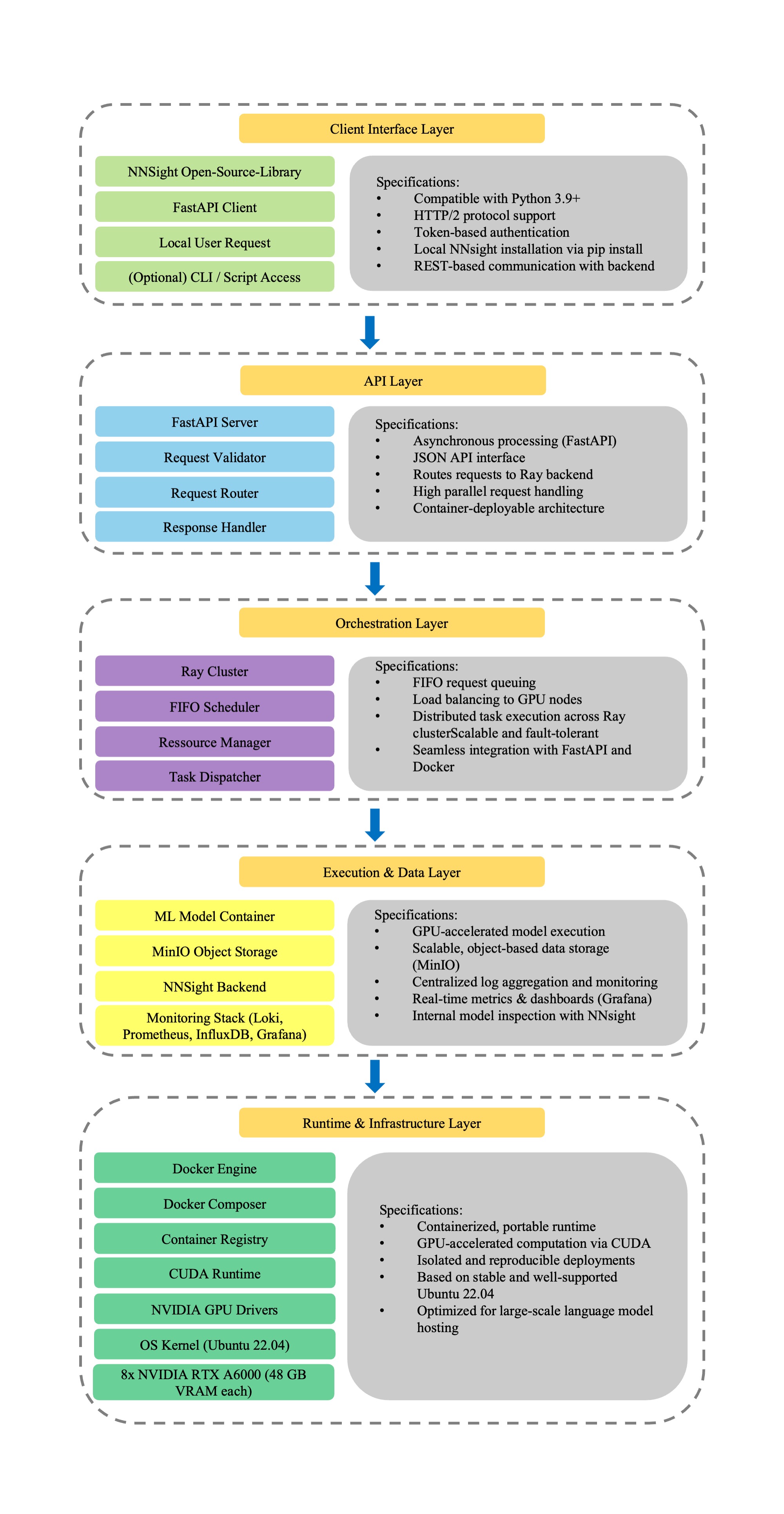}}
\caption{Layered architecture of the eDIF system with user interaction, orchestration, and GPU execution}
\label{fig1}
\end{figure}

The eDIF system employs a modular stack combining FastAPI (request interface), Ray (orchestration), and Dockerized model runtimes on NVIDIA RTX A6000 GPUs. User-side interactions are performed via NNsight, a PyTorch-based toolkit enabling deep model introspection.
Requests are issued through NNsight and routed via FastAPI to the backend, where Ray manages FIFO-based scheduling across GPU instances. Execution results are returned through the same API layer. The full stack—from user interface to physical hardware—is shown in Figure \ref{fig1}.

\begin{figure*}[hbt!]
\centering
\includegraphics[width=\textwidth]{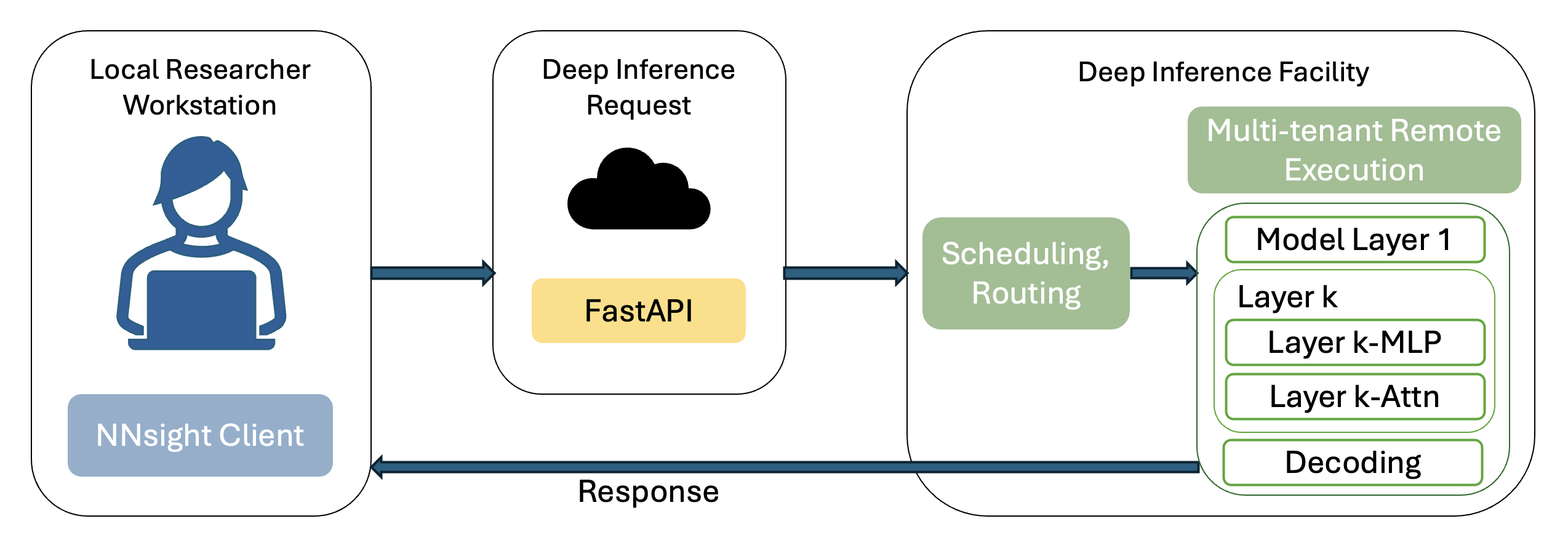}
\caption{NNsight distributed execution architecture for remote neural network inspection }
\label{fig:nnsight_architecture}
\end{figure*}

NNsight is a PyTorch-based analysis tool designed to expose and manipulate internal states of deep neural networks. It enables researchers to inspect, trace, and alter intermediate activations and parameters during model execution. In the illustrated workflow (Figure \ref{fig:nnsight_architecture}), user-defined trace functions are executed locally, while model inference is delegated to a remote backend infrastructure based on the original NDIF architecture. This hybrid setup allows researchers to inject fine-grained inspection points into remote multi-tenant inference pipelines without compromising execution efficiency. The local system (eDIF) retains architectural compatibility with NDIF, ensuring interoperability across deployments.

\subsection{External Infrastructure and HPC Integration}
\label{sec:external_hpc}

To extend computational resources beyond the dedicated eDIF server, selected external infrastructures were explored in close collaboration with partner institutions. The goal was to evaluate portability across heterogeneous systems and assess the adaptability of the platform under real-world HPC constraints.
Among these partners, Friedrich-Alexander-Universität Erlangen-Nürnberg (FAU) provided access to its AMD-based HPC cluster. This environment, equipped with eight AMD Instinct MI300X GPUs (each featuring 192 GB of HBM3 memory) enabled valuable experimentation with alternative hardware architectures and container runtimes.

The eDIF deployment on the FAU HPC cluster was implemented as an isolated instance, fully independent from the primary eDIF deployment. It included only the core services required for model inference—specifically, components such as Ray and FastAPI—while complementary modules like monitoring dashboards and visualization tools were excluded due to time and system constraints. Since the original NDIF stack relies on a Docker-based architecture, the transition to the FAU HPC environment required porting these core services to Apptainer (formerly Singularity), a container runtime compatible with non-root, multi-user HPC systems. Docker-specific features such as image builds, service orchestration, and networking had to be either manually reconfigured or replaced with lightweight shell-based alternatives where \mbox{necessary}.

Porting the system to AMD hardware also required switching from NVIDIA’s CUDA framework to ROCm (Radeon Open Compute). While most of the PyTorch-based model code remained compatible, minor adjustments, such as replacing NVIDIA's NCCL (a library for efficient multi-GPU communication) with AMD's RCCL counterpart, were sufficient to preserve the core inference workflows, thanks to PyTorch’s runtime abstraction \cite{b20, b21}.

Unlike persistently running container services, HPC platforms allocate resources on demand with fixed runtime limits. This required adapting the inference pipeline to align with scheduled execution workflows and introduced longer turnaround times during testing. In addition, the porting of containerized components from the eDIF system to the AMD-based HPC environment involved ROCm-specific adjustments and reconfiguration of existing image layers, as certain CUDA-based assumptions had to be reworked for compatibility.

Although a full-scale deployment was not pursued due to time constraints, the collaboration with FAU provided essential insights into the portability and constraints of model-serving infrastructure on shared academic compute platforms. The institutional support, offered without formal resource agreements, reflects a strong commitment to open research and inter-university cooperation.

\section{Deployment and Operation}
Beyond the setup of the technical infrastructure, the focus was placed on the concrete configuration, model deployment and stable runtime monitoring and management.
\subsection{Setup}
The system setup followed the official NDIF guide, cloned from GitHub and adapted for the eDIF context \cite{b22}. Configuration included ports, model instances, and GPU availability. Initial testing was done via API queries after container startup.
Additional adaptations included a Traefik-based reverse proxy to enable secure external access. Monitoring components (Loki, InfluxDB, Prometheus, Grafana) were configured during setup for later integration.
\subsection{Model Provisioning}
Table I lists the main LLMs that were deployed and used extensively during the eDIF project. These models remained active for most of the runtime and were selectively replaced based on specific needs or experimental requirements. Additionally, the Llama 4 Maverick model was hosted on an external FAU server to support further testing.

\par\par\par

\begin{table}[h]
\centering
\renewcommand{\arraystretch}{1.3}
\setlength{\tabcolsep}{4pt}
\begin{tabular}{|>{\centering\arraybackslash}m{2.7cm}|
                >{\centering\arraybackslash}m{2.9cm}|
                >{\centering\arraybackslash}m{2.2cm}|}
\hline
\textbf{Model} & \textbf{Model size} & \textbf{Required GPUs} \\
\hline
GPT-2 small \cite{b23} & 124 million parameters & 1× RTX A6000 \\
\hline
DeepSeek-R1-Distill-Llama-8B \cite{b24} & 8.03 billion parameters & 2× RTX A6000 \\
\hline
DeepSeek-R1-Distill-Llama-70B \cite{b25} & 70.6 billion parameters & 5× RTX A6000 \\
\hline
\end{tabular}
\vspace{2mm} 
\caption{Evaluated models}
\label{tab:modelle-spalte}
\end{table}

\vspace{-2ex} 
All models were sourced from public Hugging Face repositories and integrated without additional wrappers. Model access was provided through a seamless connection between NNsight and the eDIF backend, appearing identical to standard Hugging Face usage from the user's perspective.\par

GPU allocation for each LLM was based on expected memory requirements. This approach provided reasonable resource planning but created inefficiencies since GPUs could not be shared between models. Unused memory on individual GPUs remained inaccessible to other models, creating a gap between theoretical GPU capacity and actual utilization.

\subsection{Operation and Monitoring}
After model provisioning, the focus shifted to stable runtime operation and continuous system monitoring. The monitoring setup relies on a combination of established tools:
\begin{itemize}
    \item \textbf{Prometheus}: Time-series metric collection and storage; foundation for performance data representation. \cite{b17}
    \item \textbf{InfluxDB}: Long-term retention of historical time-series data; supplementary role alongside Prometheus. \cite{b18}
    \item \textbf{Loki}: Log aggregation via label-based indexing; no full-text indexing of log contents. \cite{b16}
    \item \textbf{Grafana}: Unified dashboard interface; visualization of metrics and logs from multiple sources. \cite{b19}
\end{itemize}
This monitoring stack provides real-time observability and plays an important role in ensuring reliable system performance.

\section{Feasibility Study Design and Methodology
}
The eDIF feasibility study was meticulously planned and executed between May 12th and June 25th, 2025, at Ansbach University of Applied Sciences. Conducted within a Master's program the primary objective of this study was to establish and refine a dedicated computational infrastructure tailored specifically for interpretability research.  Recruitment for the study via e-mail and Discord-community targeted researchers actively engaged in mechanistic interpretability research, highlighting the availability of powerful GPU resources alongside a PyTorch-based experimental API integrated seamlessly with the NNsight toolkit. The researchers were provided with free access to the high-performance infrastructur of eDIF.
    
The recruitment process resulted in 29 applications from researchers based at various institutions and universities across Europe. Selection criteria included affiliation with a European country, a clear research focus on mechanistic interpretability, and a demonstrable motivation to engage with the platform's capabilities. Many applicants also indicated plans to explore advanced interpretability techniques such as sparse autoencoder training and evaluation, circuit tracing, and approaches aimed at transforming large language models from 'black boxes' into more interpretable 'white boxes'. Additionally, the feasibility of executing the proposed experiments on the available infrastructure (e.g., model compatibility, GPU requirements) played a decisive role in participant selection. From these applicants, 16 researchers were carefully selected based on their demonstrable expertise in interpretability and their explicit need for computationally intensive tasks. Researchers from seven distinct European nations participated —including the United Kingdom, Netherlands, Spain, Italy, Germany, Austria, and France. These participants came from a broad spectrum of academic institutions and brought with them a wide range of research objectives and disciplinary perspectives.
    
The temporal structure of the feasibility study followed a clearly defined sequence, designed to support both the technical integration and the scientific rigor of the research conducted. During the initial phase, participants were onboarded through a structured process that included detailed tutorial notebooks tailored to two core use cases. One focused on investigating gender bias in LLM outputs. The other centered on protective instinct analysis in the neuron activations of LLMs, using the eDIF backend to trace and manipulate relevant computational pathways. Additionally, a dedicated onboarding notebook was provided. The latter provided a step-by-step introduction to the eDIF backend and guided users through the setup of API access, including token configuration. In the next step, the system infrastructure was finalized, and individual research objectives were aligned with the platform’s capabilities. This was followed by an extended experimental phase, during which users implemented probing, tracing, and intervention experiments within their respective use cases. A structured feedback checkpoint in the middle of the study facilitated rapid iteration and improvements based on user experience. The final stage encompassed the synthesis of empirical results, reflective analysis, and formal documentation—laying the foundation for future developments and evaluation of broader applicability.
The study's experimental protocols were structured around three primary forms of intervention: probing specific neuron activations to understand their individual roles, tracing activations across different layers of neural network models, and performing targeted interventions designed to explore causal relationships within the networks. 

\section{Results}
This section outlines the main results of the eDIF project, combining technical insights and user experiences to assess the system’s performance and practical applicability.
\subsection{Technical Findings}
The technical evaluation of the eDIF system revealed several key insights. The deployment utilized a dedicated Ubuntu server equipped with eight NVIDIA RTX A6000 GPUs. CPU and RAM requirements were minimal, with GPU capacity acting as the primary constraint. In practice, the  allocation of GPUs based on estimated memory requirements proved workable but suboptimal. The inability to share GPUs across multiple models led to inefficiencies, as memory on individual devices remained unused. This resulted in a noticeable gap between theoretical GPU availability and effective resource utilization.

Despite this, models up to the scale of DeepSeek-R1-70B could be provisioned successfully, provided adequate hardware. The official NDIF deployment guide proved transferable, allowing seamless integration with NNsight. Some runtime issues were observed when models were distributed across multiple GPUs, occasionally leading to crashes due to access to shared parameters (e.g., weights and biases).\par

The Docker-based system architecture—combining Ubuntu, Ray, FastAPI, and MinIO—proved stable in daily operation. Monitoring tools such as Prometheus, Loki, InfluxDB, and Grafana functioned reliably, enabling useful real-time diagnostics and visualization, though not essential for basic model serving.\par
The migration experiments to AMD-based systems using the FAU HPC cluster demonstrated the general feasibility of operating NDIF on ROCm instead of CUDA. PyTorch’s internal abstractions allowed reuse of much of the existing codebase with only minor modifications (e.g., NCCL → RCCL) \cite{b20, b21}.\par
However, the HPC environment posed structural challenges. NDIF’s Docker-based setup was incompatible due to root access restrictions; therefore, a partial migration to Apptainer was required. This involved extensive reconfiguration, as Apptainer lacks support for key Docker features. Persistent services were further hindered by HPC scheduling constraints and time-limited compute sessions \cite{b26}. \par

\subsection{Pilot user activity and feedback}

The two central components of the eDIF framework received notably positive feedback from participants: first, NNsight, a versatile PyTorch-based API that facilitates model interventions; and second, eDIF, the underlying infrastructure. \par
\begin{figure}[H]
    \centering
    \includegraphics[width=\linewidth]{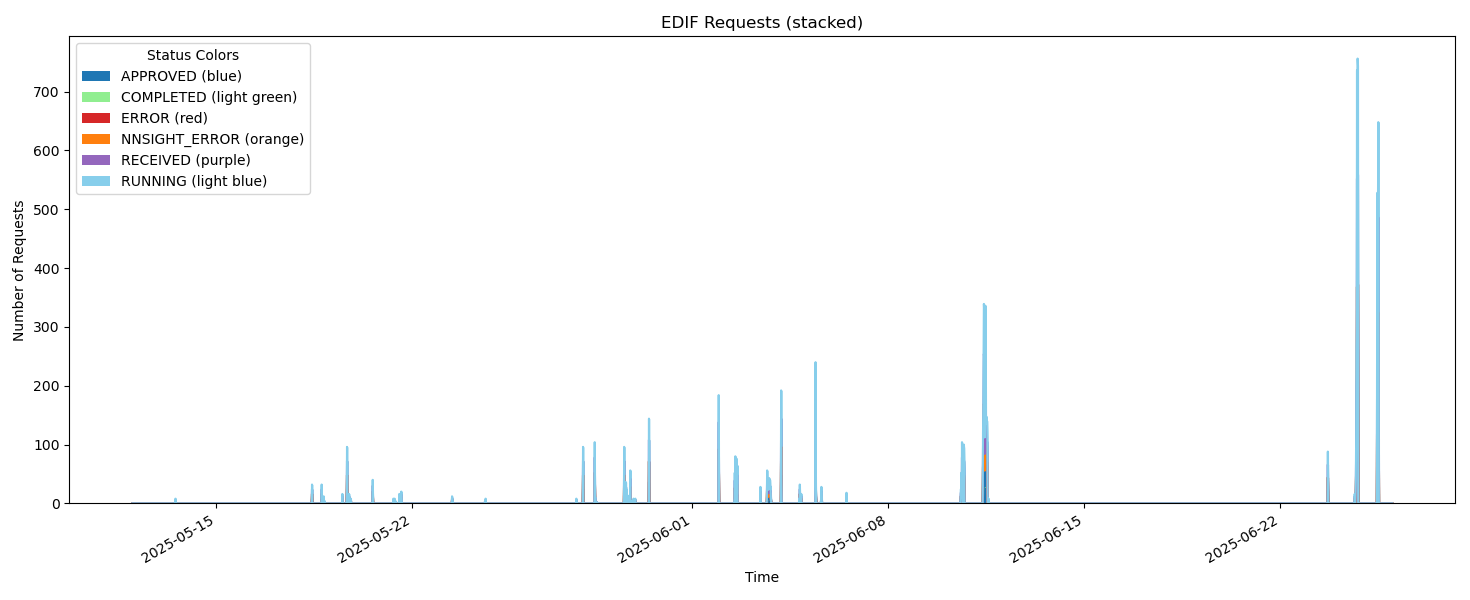}
    \caption{eDIF server activities during feasibility study}
    \label{fig:activity}
\end{figure}

    User activity throughout the study period showed a steadily increasing engagement curve, with a marked peak in request volume toward the final project week. As visualized in the platform’s internal monitoring dashboard (Figure \ref{fig:activity}), requests to the eDIF server rose from an initially sparse distribution in mid-May to over 700 in a single day by late June. Despite this sharp increase in load, the system remained functionally stable, with the majority of requests successfully processed. Intermittent model freezes and isolated error states did occur. 

\vspace{-0.5em} 

\begin{figure}[H]
    \centering
    \includegraphics[width=\linewidth]{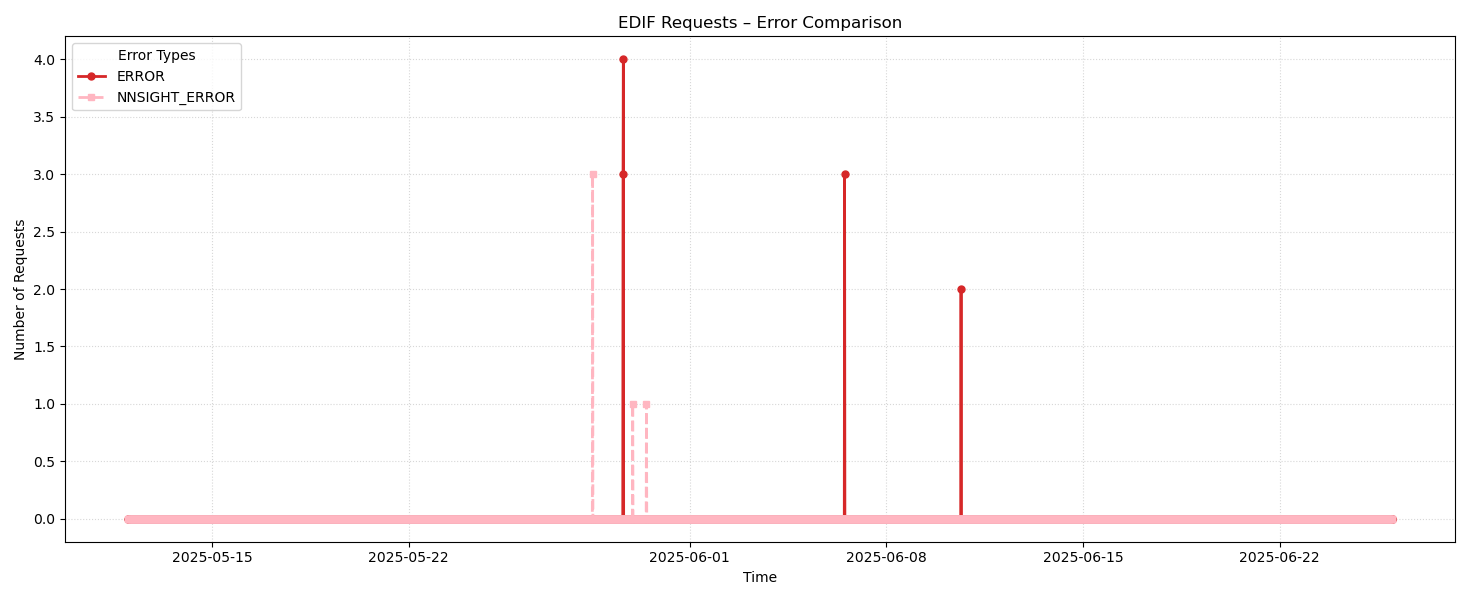}
    \caption{eDIF requests – Error comparison during feasibility study}
    \label{fig:errors}
\end{figure}

       User experience evaluations reflected high satisfaction, with mean ratings of 4.2 out of 5 for onboarding and ease of use, and 4.0 for the intuitiveness of the overall framework. Participants particularly emphasized the advantages provided by remote execution capabilities, which enabled  interventions on models without the requirement of extensive local computational infrastructure. NNsight was frequently identified as superior in comparison to existing alternatives such as TransformerLens \cite{b1}, primarily due to its remote execution functionality and seamless integration into established PyTorch workflows.

       During the study, participants actively used eDIF/NNsight for a range of tasks in the field of mechanistic interpretability. These included methods such as causal tracing, activation patching \cite{b27}, logit lens analyses \cite{b28}, and linear probe training \cite{b29}, as well as research efforts related to attribution, circuit analysis \cite{b30}, and sparse autoencoder \cite{b29} representations. Activation Patching and Logit Lens were among the most frequently used features. The predominantly accessed models included GPT-2 and DeepSeek-R1-Distill-Llama-8B, with occasional utilization of DeepSeek-R1-Distill-Llama-70B. 
       Despite its strengths, the platform encountered certain technical and usability-related challenges. Reported issues included prolonged download durations for activation data—sometimes extending up to 30 minutes—as well as intermittent server instabilities and execution interruptions (Figure \ref{fig:errors}). Users specifically requested improvements such as automated handling of connection disruptions and enhanced error management mechanisms, including retry capabilities. Although the infrastructure was generally characterized as near production-ready, bandwidth constraints and the lack of integrated support for remote dictionary learning occasionally impeded efficient workflow integration. Still, approximately half of the users estimated time savings between 10–30\% when employing eDIF, while a further 33\% experienced savings ranging from 30–50\%. One participant even reported efficiency improvements exceeding 50\%. 
       
       Feature enhancement requests predominantly centered around expanding the existing model repository to include modern, compact architectures such as Qwen and Gemma3 \cite{b31, b32}. Additionally, participants advocated for improved modularity through functional abstractions like \texttt{extract\allowbreak\_activations} expressing an interest in executing sequential processing steps (e.g. Top-k selection followed by probing) within a single operational task \cite{b1, b33}, as well as remote accessibility to analytical tools like Logit Lens. Further suggestions encompassed the development of visual aids (e.g., Top-5 Neuron views, Neuronpedia-like interfaces \cite{b34}, performance enhancements (e.g., compressed outputs and VRAM usage metadata), and comprehensive, dedicated onboarding documentation. Users consistently recommended the provision of a standalone Python package to simplify code initiation and clearer, generalized coding examples covering various model types.
       
       Regarding No-Code/Low-Code features in the future, their overall relevance received an average rating of 4 out of 5. The potential of these features to significantly lower entry barriers for non-technical team members, students, and interdisciplinary collaborators received unanimous approval, reflected in 100\% of participants rating this potential at the maximum score of 5/5. Specific functionalities requested included probing, logit lens analyses, activation patching, and representation steering. However, participants also raised critical considerations about the possible consequences of abstracting inherent complexities, noting potential impacts on transparency, reproducibility, and interpretability of results.
       
       Community-building emerged as another significant area for future development. The introduction of shared asynchronous collaboration environments, such as collaborative notebooks and open code discussions, was deemed highly valuable, with 83.4\% of participants rating this feature as 4 or 5 out of 5. Similarly, live interactive formats such as reading groups received enthusiastic support. 
       
       In summary, 83.3\% of the study participants reported partial integration of eDIF/NNsight into their research processes during the feasibility study, with unanimous interest (100\%) in sustained access to the platform within Europe. 
\section{Discussion}
The findings of the eDIF feasibility study underscore the practical viability and substantial utility of the eDIF/NNsight framework for mechanistic interpretability research across Europe. This dual validation—grounded in objective technical benchmarks and reinforced by systematic user evaluations—demonstrates the framework’s robustness and practical impact.

\subsection{Discussion of the technical results}
The system architecture implemented within the eDIF project — based on Ubuntu, NVIDIA GPUs, Docker, Ray, and FastAPI — proved to be stable and scalable. This setup provides a reliable foundation for future replication or expansion. By adopting more powerful hardware or alternative GPU types, both the number of users and model sizes could be scaled up effectively.\par
A key limitation was the absence of automatic GPU requirement estimation. Model deployments had to be tuned manually through trial-and-error, which is feasible for individual models but becomes inefficient at scale. Developing a (semi-)automated GPU profiling tool should be a priority for broader deployment scenarios.\par
Monitoring components (Prometheus, InfluxDB, Loki, Grafana) provided valuable insights into system load and performance. For future scalability, extending these tools with additional metrics could enhance debugging and optimization capabilities. Moreover, shifting from a purely pull-based approach to a push-based notification system (e.g. alerting admins in case of throttling or runtime errors) would reduce the need for manual checks and enable more proactive system management.\par
Experiments on the AMD-based HPC cluster at FAU yielded transferable insights beyond the local context. Resource allocations on HPC systems are inherently time-limited and thus incompatible with persistent APIs that require continuous model inference. The lack of Docker support — due to root permission risks — required an alternative deployment strategy via Apptainer. However, Apptainer lacks full feature parity with Docker, leading to complex migrations involving extensive rewrites of container logic, scripts, and networking setups. Furthermore, adapting CUDA-based container builds for ROCm occasionally required direct code changes or runtime modifications to existing library functions.  Technical solutions must be developed for these challenges to enable effective scaled deployment of the NDIF structure on HPC clusters.\par
Despite these constraints, the migration from CUDA to ROCm proved technically feasible. PyTorch’s internal abstractions allowed most of the codebase to be reused, requiring only minor adjustments. This finding is particularly significant as certain AMD GPUs (such as MI300X and MI325X) demonstrate competitive, and in some inference scenarios even superior, price-performance ratios compared to NVIDIA's H100 and H200. This positions AMD-based systems as a compelling option for future scaling of infrastructures such as NDIF or eDIF. \cite{b35}\par

\subsection{Discussion of the user feedback}
   The substantial increase in platform utilization toward the study's conclusion highlights a growing familiarity and comfort among participants, indicative of successful onboarding and training efforts. However, it is important to note that feedback participation was relatively low, potentially limiting the comprehensiveness of insights and generalizability of the findings. At the same time, the quality and depth of the feedback that was received proved highly valuable and highlighted specific areas of strength and targeted improvement opportunities for the platform. Additionally, phases of inactivity were observed, coinciding with European holidays and periods when users used the platform infrequently due to competing responsibilities.

While the observed surge in server requests, exceeding 700 daily requests at peak, demonstrates significant interest and operational scalability, the emergence of intermittent errors under high load conditions emphasizes the need for infrastructure refinement, particularly regarding bandwidth management and error handling protocols. From a scientific standpoint, the observed error rates fall within an acceptable threshold for a pilot-phase implementation at a smaller academic institution. As such, they should be interpreted as expected fluctuations inherent to an early-stage research infrastructure rather than as critical limitations to its exploratory utility. User feedback consistently identifies remote execution as a distinguishing advantage of the framework, suggesting a clear preference over existing alternatives. Additionally, participants expressed a clear desire for an expanded model offering, including access to even larger and more diverse language models. A frequently voiced requirement was the ability to flexibly switch between models during active sessions. This capability was seen as essential for enabling rapid experimentation and facilitating efficient comparisons within interpretability workflows. Nevertheless, the occurrence of performance bottlenecks, such as extended download durations and isolated model execution freezes, necessitates targeted technical optimizations to enhance robustness and user productivity. Further integration of automated error recovery mechanisms and improved functional modularity could substantially streamline workflows, particularly when handling sequential interpretability tasks.

Lastly, the strong interest in No-Code/Low-Code capabilities and collaborative community features suggests a strategic direction for future platform development. These enhancements promise to broaden accessibility, foster interdisciplinary collaboration, and facilitate broader adoption within diverse research communities.

\section{Future Work
}

   The eDIF feasibility study has laid a solid foundation for advancing AI interpretability infrastructure within the European research ecosystem. Based on the positive outcomes and clear user demand, the next phase will include both confirmed developments and exploratory initiatives. These focus areas address infrastructure consolidation, funding pathways, community expansion, and research capabilities. The roadmap outlined below reflects both immediate actions and strategic directions under active evaluation.

\subsection{Confirmed Developments}
Following the technical success of the MVP, long-term infrastructure support has been confirmed by the participating institutions. The current server servers with RTX 6000 and H100 at Ansbach University of Applied Sciences, equipped with NVIDIA RTX A6000 GPUs, will remain operational and continue to be available to selected pilot users beyond the official project period. In parallel, a potential expansion is being negotiated with Heilbronn University to secure access to an H200 cluster. This upgrade would significantly increase the system’s capacity for large-scale or concurrent interpretability experiments.
In addition, a deployment of AMD-based GPUs is planned to evaluate multi-vendor compatibility under the NDIF framework. This setup will enable benchmarking of NDIF performance across heterogeneous architectures. System monitoring with Prometheus and Grafana is already in place, ensuring comprehensive insights into hardware utilization and experiment throughput.
User support will also be strengthened: scheduled check-ins, live support channels, and usage logs will ensure researchers receive timely assistance and feedback. These support mechanisms are designed to improve researcher experience and foster reproducibility across experimental setups.
Initial communications with national and international stakeholders have taken place to explore potential funding opportunities. Further steps are currently under consideration. In parallel, grant proposals have been submitted for one strategic areas: the development of no-code and low-code interfaces to broaden accessibility.
Finally, community development is progressing. A European research network centered on eDIF is being prepared, with the goal of enabling long-term collaboration and shared tooling across institutions. Feedback gathered during the MVP will be formalized into onboarding improvements, training documentation, and platform tutorials and website for eDIF. Future user recruitment will target online communities such as Discord and the NNsight developer network, with the aim of reaching a broader and more diverse user base.

A dedicated platform is already accessible through the eDIF website, offering streamlined access to onboarding materials, comprehensive documentation, and supporting the development of a growing user community. \setcounter{footnote}{3}\footnote{https://www.edif.ai/}
In parallel, online community spaces (e.g., Discord, GitHub) and community events such as reading groups, workshops, or hackathons are being planned to foster engagement, shared learning, and long-term collaboration.

\subsection{Strategic Exploration and Research Directions}
Beyond operational continuity, several research-driven and usability-focused initiatives are under consideration for the coming project phase. These efforts aim to increase scientific utility, platform flexibility, and accessibility. 

In parallel, online community spaces (e.g., Discord, GitHub) and community events such as reading groups, workshops, or hackathons are being planned to foster engagement, shared learning, and long-term collaboration.

\textbf{Model and Experiment Diversity}
A key focus will be enabling cross-model interpretability studies, such as comparing the internal behavior of DeepSeek and Qwen models. These experiments are intended to explore the generalizability of interpretability methods across architectures. In addition to large-scale international models, particular attention will also be given to smaller European models that are otherwise hard to access, such as BCC/Salamander (developed at HPC Barcelona) and Teuken (developed at Fraunhofer IIS). The deployment of multimodal models (e.g., Qwen-2.5-VL) is also under evaluation, allowing for interpretability research in vision-language contexts.
Moreover, the project plans to investigate how model quantization and GPU-level sharding influence interpretability performance. This includes analyzing trade-offs between reduced resource consumption and the fidelity of activation analysis, probing accuracy, or causal tracing workflows. To support such studies, hot swapping between different model instances within the same session is a highly requested feature. This functionality would significantly enhance the efficiency of comparative experiments and aligns with future developments outlined in the broader NDIF roadmap.

\textbf{Usability and Access Expansion}
To improve researcher productivity and lower entry barriers, several platform and tooling enhancements are being considered:

\begin{itemize}
    \item Integration of visual, no-code interfaces and UI components (e.g., Transluce, Neuronpedia, activation patching, logit lens) to support intuitive and modular model exploration without requiring code-level interaction.
    \item Support for JupyterLab and VS Code to streamline local development through improved debugging capabilities, kernel compatibility, and seamless integration with existing workflows.
    \item Provision of standalone Python packages—including all required standard libraries and dependencies—for NNsight-based workflows with simplified setup.
    \item Expansion of onboarding materials, including tutorial notebooks, workflow templates, and interactive examples to accelerate familiarization and experimentation.
    
\end{itemize}
These improvements are intended to enhance the usability and accessibility of the platform for both expert users and interdisciplinary collaborators.

\textbf{Governance and Scalability}
As the platform scales, robust governance mechanisms will become essential. Authentication and access control systems are planned to support scalable and secure usage. This includes token-based authorization, user role management, and quota enforcement. Resource allocation will be governed by fair-use scheduling policies—such as time-slot reservations or per-user GPU quotas—to ensure equitable access across research teams.

In parallel, all data handling and usage tracking will be aligned with GDPR requirements, especially in multi-institutional or cross-border research contexts. This ensures that the platform remains compliant with European data protection standards while supporting transparent and auditable research workflows.
Taken together, these developments position eDIF as a sustainable and extensible infrastructure for AI interpretability research. By maintaining technical robustness, expanding access, and supporting new scientific directions, the platform will continue to contribute meaningfully to the transparency and reproducibility of AI systems in Europe.

\section{Conclusion
}
   This feasibility study demonstrates the viability and research value of a European NDIF-compatible interpretability cluster. The eDIF deployment successfully supported remote, structured interventions on large models and facilitated reproducible experimentation for a diverse group of researchers. Strong user interest, high usage volume, and positive evaluations of the NNsight interface confirm the platform’s relevance for academic interpretability workflows. While technical challenges such as model switching remain, proposed upgrades, including AMD GPU support, automated error handling, and extended model offerings, provide a clear path forward. The study establishes a solid foundation for scaling eDIF into a pan-European resource, fostering interdisciplinary collaboration and supporting the growing need for transparent, accessible AI systems. Future development will focus on usability improvements, no-code tools, and sustainable governance, positioning eDIF as a cornerstone of European AI research infrastructure. The challenges identified over the course of the project reflect common tensions that also arise in large-scale digital infrastructures. In light of growing demands for transparency, reproducibility, and equitable access to AI technologies, eDIF serves as a representative example of the infrastructural and communicative foundations that must be established to enable responsible research in Europe.

\section{Acknowledgment}

We would like to express our sincere gratitude to \mbox{Sigurd} Schacht (Ansbach University of Applied Sciences) and Carsten Lanquillon (Heilbronn University) for their academic supervision and institutional support throughout the project. We also thank the Computing Center of Friedrich-Alexander-Universität (FAU) Erlangen-Nürnberg for providing access to HPC infrastructure during the evaluation phase. Furthermore, we gratefully acknowledge the collaboration with the NDIF team in the United States, in particular David Bau (Northeastern University Boston) and Emma Bortz (Northeastern University Boston), whose technical advice and exchange greatly contributed to the success of the eDIF feasibility study.

\vspace{12pt}
\color{red}

\end{document}